%% file: main.tex
\documentclass[lettersize,journal]{IEEEtran}
\usepackage{amsmath,amsfonts,amssymb}
\usepackage{algorithmic}
\usepackage{algorithm}
\usepackage{array}
\usepackage[caption=false,font=normalsize,labelfont=sf,textfont=sf]{subfig}
\usepackage{textcomp}
\usepackage{stfloats}
\usepackage{url}
\usepackage{verbatim}
\usepackage{graphicx}
\usepackage{cite}
\usepackage{booktabs}
\usepackage{xcolor}
\usepackage[table]{xcolor}
\usepackage{pifont}

\usepackage[colorlinks,bookmarksopen,bookmarksnumbered,citecolor=blue,urlcolor=blue]{hyperref}
\input{symbol}

\renewcommand{\baselinestretch}{0.964}
\newcommand{\ms}[2]{$#1{\scriptstyle\pm #2}$}
\newcommand{\bms}[2]{$\mathbf{#1}{\scriptstyle\pm \mathbf{#2}}$}
\newcommand{\cond}[1]{{\scriptsize\textit{(#1)}}}

\begin{document}
\bstctlcite{IEEERAL_BSTcontrol}

\title{GORDON: Graph-based Object-centric Rewards \\for Decomposition of Long-Horizon Manipulation}

\author{Andrea Protopapa$^{1}$, Davide Buoso$^{1}$, Francesca Pistilli$^{1}$, Georgia Chalvatzaki$^{2, 3, 4}$ and Giuseppe Averta$^{1}$
\thanks{$^{1}$Politecnico di Torino, Turin, Italy. 
        $^{2}$ Interactive Robot Perception \& Learning (PEARL) Lab, Computer Science Dept., TU Darmstadt, Germany.
        $^{3}$ Hessian.AI, Darmstadt, Germany. 
        $^{4}$ Robotics Institute Germany (RIG).
        }%
\thanks{This study was carried out within the project FAIR - Future Artificial Intelligence Research - and received funding from the European Union Next-GenerationEU (PIANO NAZIONALE DI RIPRESA E RESILIENZA (PNRR) – MISSIONE 4 COMPONENTE 2,
INVESTIMENTO 1.3 – D.D. 1555 11/10/2022, PE00000013). This
manuscript reflects only the authors’ views and opinions, neither the European Union nor the European Commission can be considered responsible for them.}
\thanks{Contact: \texttt{andrea.protopapa@polito.it}}
}




\maketitle

\begin{abstract}
\input{sections/00.abstract}
\end{abstract}


\section{Introduction}
\label{sec:introduction}
\input{sections/01.introduction}

\section{Related Work}
\label{sec:related_works}
\input{sections/02.related_works}

\section{Method}
\label{sec:method}
\input{sections/03.method}

\section{Experiments}
\label{sec:exeriments}
\input{sections/04.experiments}

\section{Discussion}
\label{sec:discussion}
\input{sections/05.discussion}

\section{Conclusions}
\label{sec:conclusions}
\input{sections/06.conclusions}


\bibliographystyle{IEEEtran}
\bibliography{root}

\newpage

\end{document}

%% file: symbol.tex
\makeatletter
\DeclareRobustCommand\onedot{\futurelet\@let@token\@onedot}
\def\@onedot{\ifx\@let@token.\else.\null\fi\xspace}

\makeatother

%% file: sections/00.abstract.tex
Learning long-horizon manipulation skills with reinforcement learning remains
challenging due to the complexity of reward design, the limited guidance of sparse rewards, and the high cost of manual subtask annotation.
Visual demonstrations can provide supervision for reward learning, but rewards
learned from raw pixels can be brittle and sensitive to visual variation,
background appearance, and robot motion.
In this work, we propose GORDON, a graph-based object-centric reward learning framework that learns dense rewards from action-free video demonstrations. Each visual scene is represented as a graph of detected objects and spatial relations, and a graph neural network is trained in a self-supervised manner to embed these graphs into a task-aligned latent space.
To align the representation with semantic task progress, we introduce an
activity-aware weighted pooling mechanism that emphasizes task-relevant objects
while masking robot-dominated motion. 
The dense reward is then computed as distances in the learned latent space of the current state to demonstrated goal configurations, providing a measure of task progress.
In long-horizon tasks, the temporal profile of this reward
reveals stage-wise object-state transitions, enabling automatic subtask
discovery without manual segmentation. The discovered segments are then used to
train subtask-specific rewards and specialized policies that are composed
sequentially.
Experiments on seven manipulation tasks on MAGICAL and ManiSkill3 benchmarks show that our object-centric reward improves reinforcement learning in short-horizon settings and enables successful policy learning in complex long-horizon tasks through automatic decomposition, achieving an average
success rate of $74.4\%$ across the long-horizon tasks (on average $\approx +35$ p.p. vs. best learned baseline and $\approx +25$ p.p. vs. oracle).
Project page available at \href{https://andreaprotopapa.github.io/graph-reward-learning/}{https://andreaprotopapa.github.io/graph-reward-learning/}.

%% file: sections/01.introduction.tex
\IEEEPARstart{L}{ong-horizon} robotic manipulation requires reward signals that capture
semantic progress across multiple objects and ordered interaction stages.
However, manual design of such rewards for complex manipulation tasks is notoriously difficult, and sparse task
success signals provide little guidance for reinforcement learning (RL)~\cite{ng1999policy}.
A promising direction is to learn rewards from a small set of task-specific visual demonstrations that can provide supervision about what task progress looks like, while RL can use the learned reward to improve behavior through interaction~\cite{ravichandar2020recent}. In this setting, action-free video demos are especially attractive because they do not require access to
expert actions and costly annotation~\cite{eze2025learning}. 
Nevertheless, rewards learned directly from images~\cite{zakka2022xirl} can be sensitive to task-irrelevant visual factors like background/object appearance, viewpoint, and texture, as well as the visually dominant robot motion that does not relate to task progression.
In long-horizon tasks, an additional challenge is that successful execution depends on a sequence of object-state transitions rather than on the proximity to a final visual goal~\cite{pateria2021hierarchical}.
\begin{figure}[t]
    \centering
    \includegraphics[width=0.9\columnwidth]{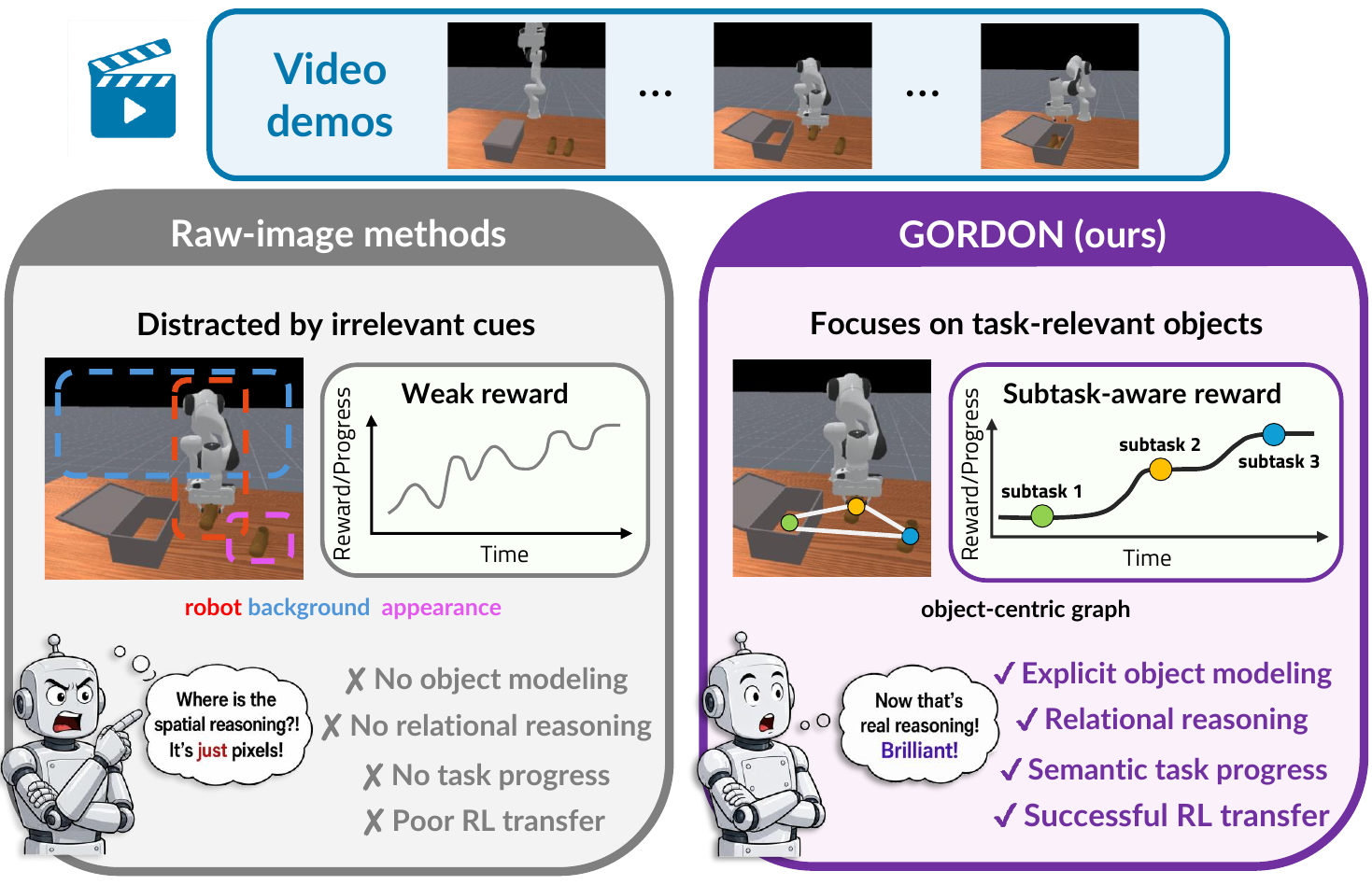}
    \vspace{-0.3cm}
    \caption{
    GORDON, our object-centric reward learning framework.
    Raw-image rewards may be distracted by spurious visual cues and transfer poorly to reinforcement learning. Conversely, using object-centric graphs, our method focuses on task-relevant objects and relations, learns a subtask-aware reward that tracks semantic progress, and transfers effectively to policy learning.
    }
    \label{fig:teaser}
    \vspace{-0.5cm}
    \end{figure}
This suggests that reward learning should focus not only on visual similarity, but also on the objects and relations whose changes define each stage of the task.
Object-centric representations offer a natural abstraction for this purpose~\cite{kumar2023graph}.
By representing a scene through detected objects and their spatial relations,
they can reduce sensitivity to irrelevant visual variation while exposing the object-state changes that underlie long-horizon progress.
Yet, object-centric reward learning still faces two key challenges: the representation must emphasize relevant objects at each stage while masking robot-dominated motion,
and the learned reward should reveal a temporal structure that supports decomposition of long-horizon tasks.

In this work, we propose GORDON, a graph-based object-centric reward learning
framework for manipulation from action-free video demos, as illustrated in Fig.~\ref{fig:teaser}.
To focus reward learning on task-relevant scene structure, each frame is converted into a graph in which nodes correspond to detected objects and edges encode spatial relations between them. A graph neural network (GNN) is trained in a self-supervised manner to embed each scene graph into a task-aware latent representation. The reward is then computed as a latent-space distance between the current scene embedding and the embedding of demonstrated goal configurations, providing dense feedback for reinforcement learning.
We further introduce an activity-aware weighted pooling mechanism that emphasizes active task objects while masking robot nodes, so that the learned reward is driven
primarily by changes in object configuration. 
Crucially, in long-horizon tasks, the temporal profile of this learned reward reveals stage-wise object-state transitions. We exploit this structure to automatically identify subtask boundaries, train subtask-specific rewards, and compose specialized subpolicies sequentially.


%
\begin{figure*}[t]
    \centering
    \includegraphics[width=\textwidth]{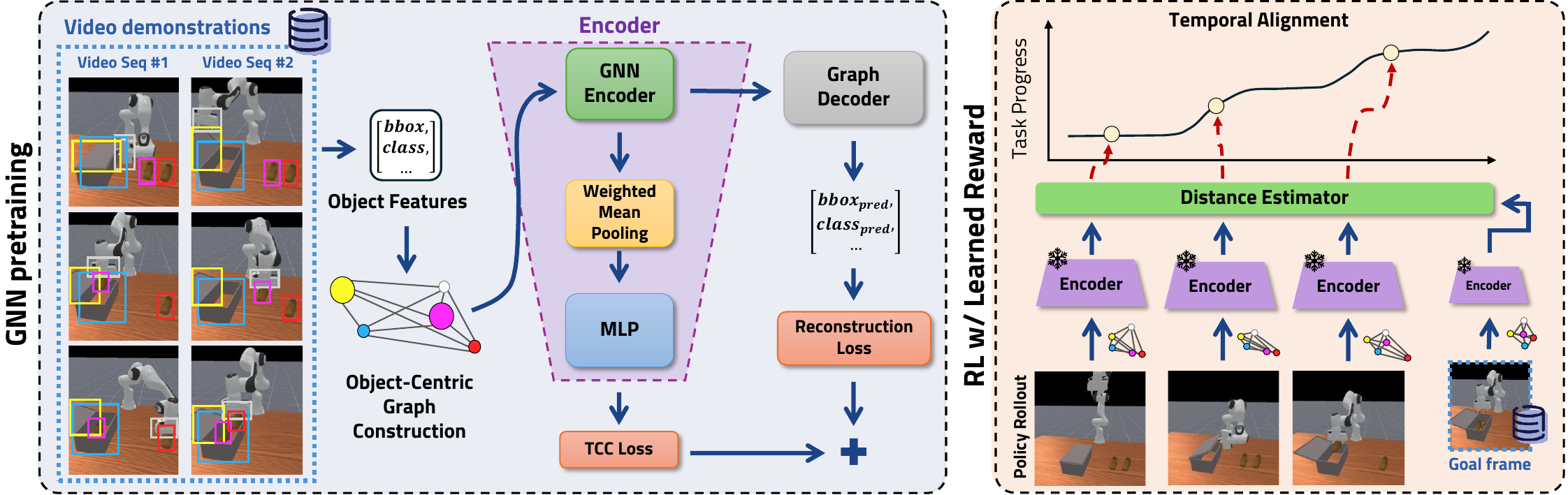}
    \vspace{-0.5cm}
    \caption{ Overview of the proposed method. Demo videos are converted into object-centric graphs and encoded with a GNN trained using temporal cycle-consistency and reconstruction losses. The frozen encoder defines a dense reward by measuring latent-space distance to the demonstration goal. This reward supports policy learning directly in short-horizon tasks and can reveal stage-wise structure for subtask discovery in long-horizon tasks. }
    \label{fig:method_overview}
    \vspace{-0.5cm}
\end{figure*}

We evaluate the method on seven manipulation tasks spanning short- and long-horizon settings. The results show that the learned graph-based reward improves downstream policy learning over pixel-based and previous object-centric baselines in short-horizon tasks. 
In long-horizon tasks, the learned reward reveals meaningful stage-wise structure, enabling automatic subtask discovery and sequential policy learning, achieving an average full-task success rate of $74.4\%$ for challenging long-horizon tasks,
$+35.3$ p.p. higher than the strongest learned or language-guided baseline and $+25.4$ p.p. w.r.t. an overly engineered manual reward serving as oracle baseline with access to privileged data.

The contributions of this work are summarized as follows:
\begin{itemize}
    \item a graph-based object-centric reward learning framework that learns
    dense rewards from action-free task demonstrations while masking
    robot-dominated motion;
    \item an automatic subtask discovery procedure and sequential policy-learning framework that  learns composable subpolicies using our dense rewards;    
    \item an extensive evaluation over seven manipulation tasks, including four long-horizon tasks, with a
    ManiSkill3~\cite{tao2025maniskill3} implementation of long-horizon benchmarks for faster, parallel and reproducible policy learning.
\end{itemize}

%% file: sections/02.related_works.tex
Reward learning from videos enables RL without manually engineered rewards or action-labeled demonstrations.
Early self-supervised methods learn temporally structured video representations for imitation and reward specification~\cite{sermanet2018time}.
Building on this, XIRL~\cite{zakka2022xirl} learns a visual embedding with temporal cycle-consistency~\cite{dwibedi2019temporal}, defining the reward as a negative distance-to-goal in latent space. Other approaches exploit large-scale human videos or pretrained visual representations to obtain general reward priors~\cite{alakuijala2023learning,ma2022vip,ma2023liv},
but are still image-based or rely on generic pretrained encoders. As a result, their rewards can be sensitive to distractors, background and visual appearance. In contrast, our goal is to learn a task-specific reward representation directly from a limited set of demonstrations, using an explicit object-centric graph abstraction. 

Object-centric reward learning addresses part of this limitation by abstracting observations into entities and relations. GraphIRL~\cite{kumar2023graph} learns rewards from object-centric video representations, improving robustness by focusing on object interactions rather than raw pixels. However, existing methods do not explicitly distinguish between objects that drive semantic task progress and motion that is only incidental to execution. This distinction is important in manipulation, where robot motion can dominate visual change while object-object relations often determine the task state. Related work has also shown that masking 
robots or task-irrelevant image regions can reduce embodiment gaps and improve visual robustness~\cite{lepert2025shadow,mirjalili2026augmented}.
Our approach complements prior work by masking robot-dominated motion not for cross-embodiment transfer, but to align rewards with semantic object-state changes via a graph representation that highlights task-relevant objects while downweighting robot nodes.

Long-horizon manipulation introduces an additional challenge, because successful behavior consists of sequential success in multiple ordered stages. Prior approaches address this through hierarchical reinforcement learning, sequential policy learning, or imitation-based segmentation and skill composition~\cite{gupta2020relay,shiarlis2018taco,chen2023sequential}.
Such methods show the benefit of decomposing long-horizon behavior, but often assume action-labeled demonstrations, task sketches, and sparse rewards.
Recent visual decomposition methods discover subgoals from phase changes in pretrained visual representations for robot learning~\cite{zhang2024universal}. While this provides an attractive off-the-shelf decomposition signal, the detected boundaries remain tied to a fixed image-level embedding space, where distance changes do not necessarily correspond to semantically meaningful subtask transitions. In contrast, we learn a task-specific reward representation directly from demonstrations in an object-centric graph space, and use its temporal profile to expose stage structure grounded in task-relevant object-state changes.

Recent methods also use language, vision-language models, or generative simulation to decompose manipulation instructions, generate intermediate rewards, or construct reusable skills for long-horizon tasks~\cite{ahn2022can,wang2024robogen,chen2025robohorizon,chen2026deco}. 
Other reward-learning methods explicitly assume demonstrations that are already segmented into subtasks using human annotations, code snippets, or VLM-generated segmentations~\cite{kim2025reds}.
Although effective, these methods rely on additional language priors, generated supervision, task descriptions, pretrained vision-language models, or predefined decomposition structure.
Conversely, we assume unsegmented action-free video demonstrations, learn an object-centric reward representation from them, and use the learned reward profile to identify candidate subtask boundaries. The resulting segments define subtask-specific rewards and policies, enabling sequential long-horizon policy learning without the need for manual substep annotation.

%% file: sections/03.method.tex
We propose GORDON (Graph-based Object-centric Rewards for DecompositiON), a task-specific object-centric reward learning framework for robotic manipulation. 
As illustrated in Fig.~\ref{fig:method_overview}, GORDON converts action-free video demonstrations into scene graphs, embeds them with a GNN into a task-aware latent representation, and derives dense rewards by comparing current scene embeddings with demonstrated goal embeddings.
This reward can be used directly for short-horizon policy learning and, in long-horizon tasks, its temporal profile is used to discover subtask structure.

\textbf{Problem Formulation.}
We consider the problem of learning reward functions for robotic manipulation from action-free video demonstrations. Let $\mathcal{D} = \{V_k\}_{k=1}^K$ denote a set of demonstration videos for a task $\mathcal{T}$, where each video $V_k = \{I_k^1, I_k^2, \dots, I_k^{T_k}\}$ is a sequence of image frames.

Our goal is to learn a reward function $r(s)$ that reflects task progress and can be used for downstream policy optimization with RL. 
To this end, we learn an embedding function $\phi(\cdot)$ that maps each observation to a latent space where distances to demonstrated goal-state embeddings reflect progress toward task completion.

\subsection{Graph-Based Object-Centric Reward Learning}
\label{sec:pretraining}

For each frame $I_t$ of a demonstration, we extract a set of $N_t$ object instances from detector outputs or simulator annotations,
following prior object-centric reward learning approaches~\cite{kumar2023graph}. We then represent the scene as a fully connected graph $G_t = (V_t, E_t)$ where each node $v_i \in V_t$ corresponds to a detected object and is described by the semantic class and bounding-box geometry  in 2D or 3D, according to the environment.
Edges $e_{ij} \in E_t$ encode pairwise spatial relations between objects, represented by Euclidean distances between their bounding boxes.
Unlike object-centric baselines that operate directly on bounding-box feature vectors, this graph representation allows for modeling relational structure among objects before computing the reward.

Given a sequence of graphs \(\{G_1,\dots,G_T\}\), we learn a graph
embedding function \(\phi(\cdot)\) that maps each scene graph $G_t$ to a compact
task-state representation $z_t$. We define \(\phi\) as the composition of a GNN
encoder, a permutation-invariant weighted pooling operator, and an MLP
projection head. For each graph
\(G_t\), the GNN processes node and edge features through message passing and
produces node embeddings \(\{h_{t,i}\}_{i=1}^{N_t}\), where \(i\) indexes the
detected objects in frame \(t\). These node embeddings are then aggregated into a graph-level pooled representation
\begin{equation}
\bar{h}_t =
\frac{\sum_{i=1}^{N_t} w_{t,i} h_{t,i}}
{\sum_{i=1}^{N_t} w_{t,i}} .
\label{eq:weighted_pooling}
\end{equation}
where \(w_{t,i}\) denotes the pooling weight of object \(i\) at time \(t\),
defined below.
The final graph embedding \(z_t=\phi(G_t)\) is obtained from \(\bar{h}_t\) by applying a MLP projection head.

\noindent\textbf{Weighted Graph Pooling.}
To make the representation more sensitive to semantic task progression, the pooling weights are defined as
\begin{equation}
w_{t,i} =
\left(1 + (\alpha - 1)\cdot \mathrm{active}_{t,i}\right)
\left(1-\mathrm{robot}_{t,i}\right),
\label{eq:node_weight}
\end{equation}
where \(\mathrm{active}_{t,i}\in\{0,1\}\) indicates whether object \(i\)
is active at time \(t\), \(\mathrm{robot}_{t,i}\in\{0,1\}\) indicates whether
the node corresponds to the robot, and \(\alpha\) controls the relative weight
assigned to active non-robot objects.

The activity indicator is computed from temporal displacement of bounding-box
centers:
\begin{equation}
\mathrm{active}_{t,i} =
\mathbb{I}
\left(
\frac{\|c_{t,i}-c_{t-k,i}\|_2}{\mathrm{diag}_{t,i}}
>
\tau
\right),
\label{eq:activity}
\end{equation}
where \(c_{t,i}\) is the center of the bounding box of object \(i\) at time
\(t\), \(\mathrm{diag}_{t,i}\) is its bounding-box diagonal, \(k\) is the
temporal window, and \(\tau\) is an activity threshold. Normalizing by the box diagonal makes the criterion scale-invariant. In practice, we smooth activity estimates over time to reduce sensitivity to detector jitter, and once an object becomes active, we keep it active for the rest of the episode to preserve information about objects involved in completed stages. This persistence reflects the intuition that objects involved in earlier stages often remain relevant for representing completed progress.

The robot indicator is obtained from the object class label provided by the detector. 
Robot nodes are masked before message passing by zeroing their features and removing their incident edges, and are assigned zero pooling weight. Consequently, the graph representation is determined by non-robot objects and their relations, which are upweighted to emphasize entities whose state changes indicate task progress.

\noindent\textbf{Self-Supervised Pretraining.}
The graph encoder is trained with two complementary self-supervised objectives: \emph{temporal alignment} and \emph{structural reconstruction}. The temporal objective encourages embeddings corresponding to similar stages of the task to be close across demonstrations, even when the demonstrations differ in timing or execution details. We use temporal cycle consistency (TCC)~\cite{dwibedi2019temporal} for this purpose, encouraging the representation to capture task progression rather than raw frame similarity.

To preserve object-level information, we also use a reconstruction objective that encourages the embedding to retain descriptors such as object classes and bounding-box geometry. Let \(\psi(\cdot)\) be a lightweight attention-based object decoder inspired by~\cite{im2024egtr} that predicts object bounding-box coordinates and class logits from the latent graph representation. We train the decoder with a reconstruction loss that combines bounding-box regression and object-class prediction:
\begin{equation}
\mathcal{L}_{rec}
=
\lambda_{box}(\mathcal{L}_{1}+\mathcal{L}_{cent})
+
\lambda_{giou}\mathcal{L}_{giou}
+
\lambda_{cls}\mathcal{L}_{cls},
\end{equation}
where the terms penalize bounding-box coordinate, centroid, generalized-IoU, and object-class prediction errors, respectively. The final training objective is
\begin{equation} 
\mathcal{L} 
= \mathcal{L}_{tcc} 
+ 
\lambda_{rec} \mathcal{L}_{rec}, \label{eq:pretraining_loss}
\end{equation}
where \(\lambda_{rec}\) balances temporal alignment and reconstruction. This objective encourages the latent representation to preserve task-relevant scene structure while organizing demonstrations according to temporal progress.

\subsection{Reinforcement Learning with Learned Rewards}
\label{sec:rl_training}
After pretraining the graph encoder, we construct a dense reward in the learned latent space. 
Following prior embedding-based reward learning methods~\cite{zakka2022xirl,kumar2023graph},
we compute the goal embedding from the final frames of the demonstration
trajectories. Specifically, encoding the final graph of each demonstration, we define the average goal embedding as
\begin{equation}
z_g^{full} 
= 
\frac{1}{K} \sum_{k=1}^{K} \phi(G_k^{T_k}), 
\label{eq:full_goal_embedding}
\end{equation} 
where $G_k^{T_k}$ is the graph extracted from the final frame of demonstration $k$ with length $T_k$. Given the current observation graph $G_t$ encoded as $z_t$, the dense full-task reward is then defined as the negative distance to the average goal embedding:

\begin{equation}
\tilde{r}^{full}_t = -\|z_t-z_g^{full}\|_2 . \label{eq:full_task_reward} 
\end{equation}
This reward assigns higher values to states whose object-centric configurations are closer to the demonstrated goal in the learned task-aligned embedding space.

Although this distance-based reward provides useful shaping, distance alone can be ambiguous near task completion, especially in long-horizon manipulation. Several states may be close to the goal in latent space while still failing to satisfy the semantic success condition. 
To improve precision near terminal states, we train on top of the frozen graph embeddings a lightweight MLP classifier $c(\cdot)$ which outputs a goal probability $p_t=\sigma(c(z_t))$.
After temporal alignment, frames in the final $q\%$ of each demonstration are labeled as positive goal states, rather than using only the last frame, to capture small variations among valid terminal configurations and provide more stable classifier supervision. The remaining frames are
treated as negative examples.
At deployment, the classifier acts as a
conservative success detector and provides an additional terminal bonus only when $p_t \geq \tau_c$, where $\tau_c$ is selected on a validation split.

The final learned reward used for policy optimization is 
\begin{equation} 
r^{full}_t
= 
\tilde{r}^{full}_t
+ \beta \cdot \mathbb{I}[p_t \geq \tau_c], 
\label{eq:final_full_reward}
\end{equation} 
where $\beta$ controls the terminal bonus. The dense term provides smooth progress shaping throughout the episode, while the classifier-gated terminal bonus improves precision near true completion states. For short-horizon tasks, this 
full-task reward can be used directly to train a policy with RL.
\begin{figure}[t]
    \centering
     \includegraphics[width=\linewidth]{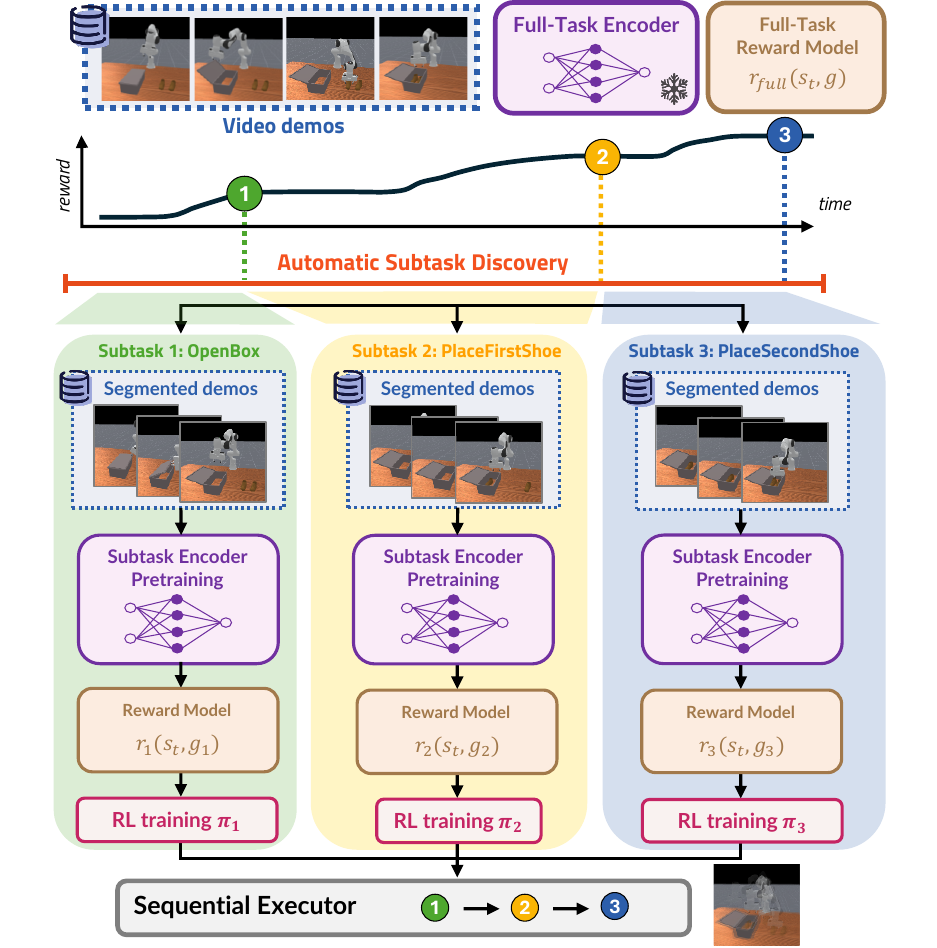}
    \vspace{-0.5cm}
    \caption{ Automatic subtask discovery and sequential policy training for long-horizon tasks, using \textsc{PutShoesInBox} as example task. The learned full-task reward is used to identify stage transitions and segment demonstrations into subtasks. Each segment is used to train a subtask-specific reward and policy, which are composed sequentially at deployment according to task progress. }
    \label{fig:automatic_subtasks_policies}
    \vspace{-0.5cm}
\end{figure}
\subsection{Automatic Subtask Discovery and Policy Composition}
\label{sec:subtask_discovery}
In long-horizon tasks, the learned full-task reward exhibits a stage-wise temporal profile. Rather than changing smoothly at every frame, the reward tends to increase around semantically meaningful transitions in the scene, such as changes in object placement, containment, or interaction state. This structure provides a useful signal for decomposing long-horizon tasks into ordered manipulation phases.

\noindent\textbf{Subtasks discovery.}
After training the full-task reward model, we temporally normalize each demonstration and analyze the temporal evolution of the predicted reward. Candidate transition frames are identified from significant changes in the reward gradient, which correspond to substantial changes in the underlying object-centric scene configuration. 
The number of subtasks is not fixed a priori, but it is instead determined by the number of transition points detected from the reward-gradient profile.
If \(M-1\) transition points are detected, the demonstration is split into \(M\) ordered segments. The resulting segments define subtask-specific datasets, each associated with a distinct phase of the overall task.

\noindent\textbf{Subtask reward and policy learning.}
For each discovered segment \(m \in \{1,\dots,M\}\), we train a subtask-specific encoder and reward model using only the corresponding demonstration segments. The local goal embedding is computed in the same way as in Eq.~\ref{eq:full_goal_embedding}, but using the final frames of the segments assigned to subtask $m$ rather than the final frames of the complete demonstrations, thus defining a local reward $r^{(m)}$. Training separate reward models for each segment is important because the full-task reward is aligned with the final task outcome and may provide weak or ambiguous feedback within an individual phase. In contrast, a subtask-specific reward tends to be more discriminative, emphasizing the object relations that should change during the current phase while treating relevant objects of other phases as context.

\noindent\textbf{Subpolicy training and composition.}
We finally optimize a dedicated policy $\pi^{(m)}$ using the corresponding subtask reward $r^{(m)}$ for each subtask $m$. This produces specialized policies for individual manipulation phases, rather than requiring a single policy to solve the entire long-horizon task from sparse or ambiguous feedback.
At deployment time, a sequential executor switches from policy
$\pi^{(m)}$ to $\pi^{(m+1)}$ when the current subtask is detected as complete.
To reduce mismatch between training and execution, each subpolicy is trained from perturbed initial states around plausible configurations produced by the preceding phase. Between consecutive subtasks, collision-aware motion planning is used only to reposition the robot arm to the initial configuration of the next subpolicy.
The resulting procedure is depicted in Fig.~\ref{fig:automatic_subtasks_policies}. 

%% file: sections/04.experiments.tex
The experimental evaluation is organized in two parts. First, we present the main results, showing that the proposed method (i) automatically recovers meaningful subtask structure from demonstrations through the learned reward profile and (ii) transfers the learned reward to downstream policy training, achieving strong success rates across both short- and long-horizon manipulation tasks. Second, we provide ablation studies that isolate the role of the main design choices of the method and analyze their effect. 

\begin{figure}[t]
    \centering
    \includegraphics[width=0.99\linewidth]{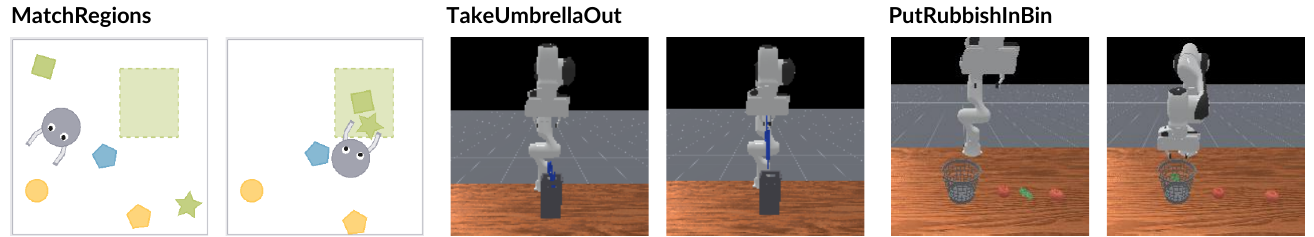}
    \vspace{-0.5cm}
    \caption{
    Short-horizon tasks require relatively compact sequences of object interactions. We use these tasks to assess whether the learned object-centric reward transfers effectively to policy optimization in simpler settings.
    }
    \vspace{-0.2cm}
    \label{fig:tasks_overview_short}
\end{figure}

\begin{figure}[t]
    \centering
    \includegraphics[width=0.99\linewidth]{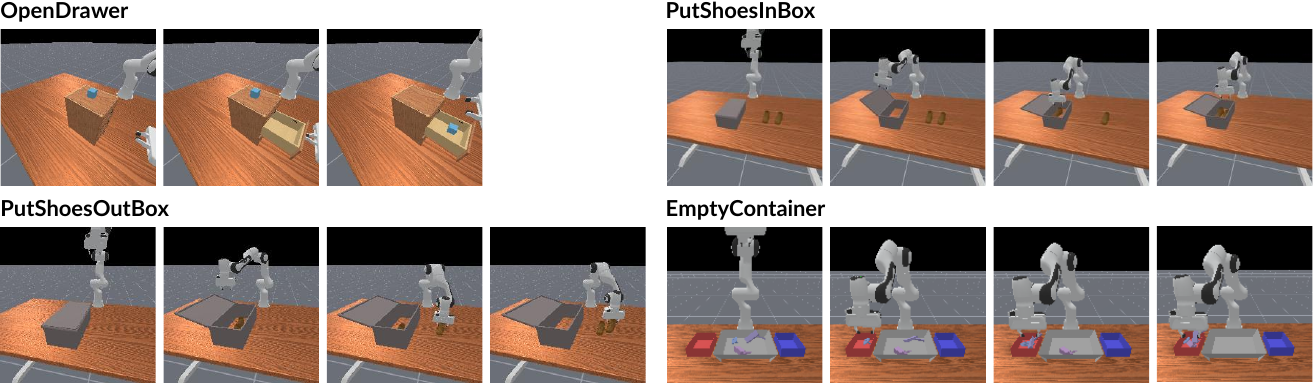}
    \vspace{-0.5cm}
    \caption{
    Long-horizon tasks require multiple ordered object-state transitions. These are used to evaluate whether the learned reward captures semantic task progress, reveals subtask structure, and supports sequential policy learning in more complex 3D manipulation scenarios.
    }
    \label{fig:tasks_overview_long}
    \vspace{-0.5cm}
\end{figure}

We evaluate our method on seven manipulation tasks spanning different levels of
difficulty, from a controlled 2D benchmark to 3D short- and long-horizon
manipulation. \textsc{MatchRegions}, from MAGICAL~\cite{toyer2020magical}, is
a short-horizon 2D task that requires placing objects into corresponding target
regions, and is used as a controlled benchmark for reward learning and
downstream RL. The other six tasks are implemented in ManiSkill3~\cite{tao2025maniskill3}
and are inspired by RLBench~\cite{james2020rlbench}. These 3D tasks include
two short-horizon tasks with a single semantic objective and four long-horizon tasks that require completing multiple ordered subtasks. They preserve the high-level
objectives and subtask structure of the original RLBench tasks while enabling
faster parallelized RL training. Figures~\ref{fig:tasks_overview_short}--\ref{fig:tasks_overview_long}
provide a visual overview of the seven tasks considered in our evaluation.

We compare against hand-crafted environmental rewards and learned-reward baselines. The environmental reward uses a privileged simulator state, therefore we refer to this as an oracle baseline for task-specific reward design. For long-horizon tasks, we evaluate two oracle variants: a single full-task environmental reward and a decomposed environmental-reward variant in which each manually specified subtask is assigned its own dense environmental reward, and the resulting subpolicies are composed with the same sequential executor used by our method. This second variant estimates the performance achievable when the decomposition and dense subtask rewards are manually engineered.
XIRL~\cite{zakka2022xirl} learns a pixel-based temporally aligned embedding from demonstration videos using temporal cycle consistency and defines reward as the negative distance to a goal embedding. GraphIRL~\cite{kumar2023graph} uses a lightweight object-centric abstraction based on detected bounding boxes and pairwise geometric relations to learn temporally aligned rewards. 
Compared with our method, it provides a simpler feature-level object representation and does not use a task-aware GNN encoder, weighted graph pooling, or robot-motion masking.
For long-horizon tasks, we additionally compare against RoboHorizon~\cite{chen2025robohorizon}, which uses language-guided task decomposition and LLM-generated staged rewards within a multi-view model-based RL framework.

\begin{figure*}[t]
    \centering
    \includegraphics[width=\textwidth]{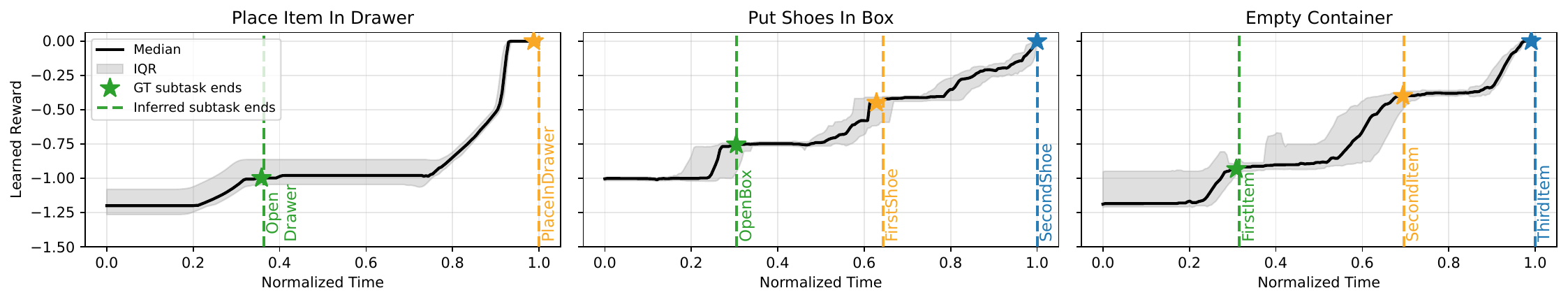}
    \vspace{-20pt}
    \caption{
    Learned full-task reward profiles on three representative
    long-horizon tasks. Stars denote simulator-annotated subtask endings, while dashed lines denote boundaries inferred from the learned reward profile.
    Their close alignment shows that stage transitions can be recovered from unsegmented demonstrations.
    }
    \label{fig:learned_reward_profiles}
    \vspace{-10pt}
\end{figure*}
All RL policies are trained with Soft Actor-Critic (SAC)~\cite{haarnoja2018soft} using the same policy architecture, action space, and training budget. Results are averaged over five random seeds. 
Reward models are trained from action-free demonstration videos collected by teleoperation in simulation. We use 100 demonstrations for \textsc{MatchRegions} and 25 demonstrations for each ManiSkill3 task. Object detections are obtained from simulator annotations, providing 2D bounding boxes for \textsc{MatchRegions} and 3D bounding boxes for ManiSkill3 tasks, semantic class labels, and robot labels for masking. These annotations are used only to construct the reward representation; RL policies are trained with the same observation space across methods.
Throughout our experiments, we set \(\alpha=2.0\), \(k=40\), \(\tau=0.005\), \(\lambda_{rec}=0.5\), and terminal bonus weight \(\beta=5\). For reconstruction, we use \(\lambda_{box}=0.9\), \(\lambda_{giou}=0.1\), and \(\lambda_{cls}=1.0\). The classifier threshold \(\tau_c\) is selected per task on a validation split, while the positive goal window is set to $q=3\%$.
For our method, object-centric graphs are encoded with a GraphTransformer encoder~\cite{yun2019graph}, which is pretrained with the objectives described in Section~\ref{sec:method} and then frozen during policy optimization. For short-horizon tasks, a single full-task reward is used for RL training. For long-horizon tasks, we first train a full-task reward model, segment demonstrations using its reward profile, and then train one subtask-specific reward model and RL policy per discovered segment. 

For each seed, success is measured over 16 evaluation episodes using task-specific simulator-state success predicates. For long-horizon tasks, we additionally report cumulative subtask success rates in Tab.~\ref{tab:sequential_merging_analysis}. 
For methods with explicit subpolicies, such as our method and the decomposed environmental-reward oracle, we evaluate the sequentially composed policy. For the single-policy baselines, we execute the full-task policy and retrospectively measure how many ordered subtask predicates are satisfied. Thus, a rollout is counted as successful for $S_1 \rightarrow S_2$ only if both subgoals are completed in order. 
RoboHorizon is included only in the full-task comparison because we use the published full-task success rates and do not have access to rollout trajectories for computing the same subtask predicates. 
The comparison should therefore be interpreted as a semantically matched benchmark comparison. We further evaluate the correspondence between the original RLBench task and our ManiSkill3 reimplementation under matched training configurations in Section~\ref{sec:rlbench_maniskill}.

\subsection{Main Results}
\label{sec:main_results}
    
\input{tables/full_task_sr_final}

\noindent\textbf{Learned Reward Profiles.}
\label{sec:learned_reward_profiles}
We show representative full-task reward
profiles for three of the four long-horizon tasks in Fig.~\ref{fig:learned_reward_profiles}. The goal of this analysis is
to verify whether the reward learned from full-task demonstrations contains an
interpretable temporal structure. The reward exhibits stage-like
transitions that closely align with simulator-annotated subtask endings.
Moreover, the boundaries inferred from the reward profile largely coincide with these annotations. Since subtask labels are not used during reward learning, this result shows that candidate subtask boundaries can be recovered directly from the learned full-task reward and used to define simpler subtask-specific policy-learning problems for long-horizon policy training.

\noindent\textbf{Policy Evaluation Across Tasks.}
\label{sec:policy_success_rate_results}
We now discuss the policy evaluation results. Tab.~\ref{tab:full_task_success_all}
reports the full-task success rate for all seven tasks. For long-horizon tasks,
Tab.~\ref{tab:sequential_merging_analysis} additionally reports cumulative
subtask success rates.
For decomposed methods, these are obtained by composing the learned subpolicies sequentially; for single-policy baselines, we execute the full-task policy and measure which ordered subtask predicates are satisfied during the rollout. 
Values in parentheses indicate the conditional success rate given completion of the previous subtask.
The results in Tab.~\ref{tab:full_task_success_all} show that the learned reward transfers effectively to policy training. 
On short-horizon and single-stage tasks, our method achieves strong performance, comparable with the environmental oracle policy. 
The main gains appear in long-horizon tasks, where the proposed reward decomposition and sequential policy training strategy substantially improve full-task completion. 
Across the four long-horizon tasks, our method achieves an average success rate of $74.4\%$, improving over RoboHorizon by $35.3$ p.p. and over the decomposed environmental-reward reference by $25.4$ p.p., while the single full-task environmental reward, XIRL, and GraphIRL fail to complete the full task.
Tab.~\ref{tab:sequential_merging_analysis} provides a more detailed view of these long-horizon failures. 
The single full-task environmental reward and learned baselines are not always uninformative. In several cases, they complete the first stage of the task, such as opening the drawer or opening the box. 
However, their performance drops sharply on later cumulative stages, indicating that their reward signals can induce partial progress but do not provide sufficient stage-wise guidance for full long-horizon completion. 
In contrast, our method maintains high conditional success across later stages, showing that the discovered subtask structure provides more reliable supervision for sequential policy learning.
The remaining failures mainly occur in the most demanding long-horizon tasks. 
\textsc{PutShoesOutBox} requires precise placement of objects outside the container, where the box no longer constrains the final object configuration, while \textsc{EmptyContainer} requires reliable sequential execution over multiple placed objects. 
Despite these challenges, our method obtains the best full-task success on all long-horizon tasks and the highest long-horizon average. 
The comparison with RoboHorizon further suggests that learning task progress directly from object-centric demonstrations can provide a precise and transferable training signal without relying on externally specified reward stages or language-generated decomposition.

\input{tables/sequential_merging_sr_final}

\subsection{Analysis and Ablation Studies}
\label{sec:ablations}

We analyze the main components of the method on \textsc{PutShoesInBox}. 
This task is long-horizon, three-dimensional, and naturally decomposes into ordered subtasks, making it an informative setting for evaluating reward discrimination, robot-motion masking, activity-aware pooling, training objectives, and subtask-level policy composition. 

\noindent\textbf{Reward Discrimination.}
\label{sec:reward_discrimination_shoes}
We first evaluate whether the learned reward assigns meaningfully different cumulative returns to successful and unsuccessful validation trajectories. We measure reward discrimination using the ratio
\[
\rho = \frac{\bar{R}^{+}}{\bar{R}^{-}},
\]
where $\bar{R}^{+}$ and $\bar{R}^{-}$ are the mean cumulative rewards over successful and unsuccessful validation trajectories, respectively. 
Since the reward is defined as a negative latent distance, successful trajectories should accumulate rewards closer to zero than unsuccessful ones. Under this reward convention, lower values of $\rho$ correspond to a larger separation between successful and unsuccessful executions.

Tab.~\ref{tab:shoes_ratio} compares reward discrimination across a pixel-based reward baseline, an object-centric baseline, and variants of our method. 
XIRL provides the image-based reference, while GraphIRL provides the closest object-centric learned-reward baseline.  
Because the original GraphIRL representation includes the robot as an entity in the scene, we additionally evaluate a variant in which robot entities are removed from its input representation. 
This allows us to test whether improved discrimination is explained only from robot masking or from the overall graph representation.
Our robot-masked model achieves the strongest discrimination, while our robot-included variant still outperforms GraphIRL
with robot nodes removed, indicating that the gain is not due only to robot masking. 

We further evaluate two robustness-oriented settings. 
First, to test sensitivity to visual appearance, we render validation trajectories with altered scene and object colors while keeping the trained reward models fixed. 
This appearance shift substantially degrades XIRL, while object-centric methods are unaffected (as expected and desired) by changes in visual appearance, because their rewards are computed from object identities and geometric relations rather than raw pixels. This supports robustness to the appearance of object-centric reward learning methods.
Second, to evaluate sensitivity to imperfect object localization, we perturb the detected bounding boxes at every frame with Gaussian center noise ($\sigma=1$cm per axis) and scale noise on the half-extents ($\sigma=5\%$ per axis) before computing graph-based rewards.
The final version of our method remains stable under this moderate localization noise, while the variant that keeps robot entities becomes less discriminative. 
Therefore, masking robot-dominated motion not only improves semantic alignment but also reduces sensitivity to noisy robot geometry in the reward representation.

\input{tables/reward_discrimination}

\noindent\textbf{Subtask-Level Reward Transfer.}
\label{sec:subtask_reward_transfer}
We study how the learned reward transfers to the subtasks discovered in \textsc{PutShoesInBox}. 
Tab.~\ref{tab:multi_task_success} reports success rates for individual subtask policies, while the composed-policy column reports sequential composition of the corresponding subtask policies.
The comparison separates the effect of the dense learned rewards from terminal bonus signals used to stabilize policy learning. 
We include an environmental bonus as an oracle diagnostic and the learned classifier-gated bonus from Eq.~\ref{eq:final_full_reward} as the fully learned variant used in our final method.
The learned reward alone solves the opening subtask but is insufficient for the contact-rich shoe-placement stages. Adding a terminal bonus substantially improves subtask learning.
Among the tested variants, the learned classifier-gated bonus yields the best composed-policy performance, showing that the dense reward provides useful shaping while the learned bonus is important for precise subtask completion and
sequential execution.
\input{tables/ablation_bonus}

\noindent\textbf{Training Objective Ablation.}
\label{sec:training_objective_ablation}
We ablate the representation-learning objectives using the reward-discrimination ratio $\rho$ defined above. 
All variants use the same graph encoder, pooling strategy, and validation trajectory split, isolating the effect of the training losses. 
Tab.~\ref{tab:ablation_losses} shows that using only temporal cycle-consistency or only reconstruction degrades reward discrimination, while combining both objectives yields the strongest separation between successful and unsuccessful trajectories. 
This suggests that temporal alignment and object-level reconstruction provide complementary supervision: the former organizes embeddings by task progress, while the latter preserves the scene structure needed for meaningful reward computation.
\input{tables/ablation_loss_shoes}

\noindent\textbf{RLBench vs. ManiSkill.}
\label{sec:rlbench_maniskill}
To assess the correspondence between the original RLBench setting and our ManiSkill3 reimplementation, we compare the two implementations on \textsc{TakeUmbrellaOut} as a controlled diagnostic for the reimplementation under matched learning settings.
Both implementations reach $100\%$ final success rate. 
However, the RLBench version requires $9.0 \cdot 10^5$ steps and $7$h $54$m, while the ManiSkill3 version reaches the same performance with $1.5 \cdot 10^5$ steps in $57$m. 
Using $32$ parallel ManiSkill3 environments, training is
further reduced to $5.0 \cdot 10^4$ steps and $27$m. 
Although limited to one task, this diagnostic supports the use of ManiSkill3 as a faster benchmark for the broader evaluation.

%% file: tables/full_task_sr_final.tex
\begin{table*}[t]
\centering
\caption{
Full-task success rate (\%). Values are mean $\pm$ std over five seeds; $^\dagger$ indicates RoboHorizon results reported in~\cite{chen2025robohorizon}. \\ Green $\Delta$ denotes average long-horizon improvement over the strongest alternative.
}
\vspace{-0.3cm}
\label{tab:full_task_success_all}
\setlength{\tabcolsep}{3.8pt}
\renewcommand{\arraystretch}{1.08}
\resizebox{\textwidth}{!}{%
\begin{tabular}{lccc|ccccc}
\toprule
Method
& \multicolumn{3}{c|}{Short-horizon}
& \multicolumn{5}{c}{Long-horizon} \\
\cmidrule(lr){2-4}
\cmidrule(lr){5-9}
& MatchRegions
& TakeUmbrellaOut
& PutRubbishInBin
& PutItemInDrawer
& PutShoesInBox
& PutShoesOutBox
& EmptyContainer
& Avg. LH\\
\midrule

\rowcolor{gray!10}
Env. Rew.
& \ms{81.3}{27.3}
& \bms{100.0}{0.0}
& \bms{100.0}{0.0}
& \ms{0.0}{0.0}
& \ms{0.0}{0.0}
& \ms{0.0}{0.0}
& \ms{0.0}{0.0}
& 0.0 \\

\rowcolor{gray!10}
Env. Rew. (decomp.)
& -
& -
& -
& \ms{62.5}{39.9}
& \ms{68.7}{39.0}
& \ms{41.2}{39.9}
& \ms{23.8}{12.0}
& 49.0 \\

\midrule

XIRL
& \ms{14.3}{20.3}
& \ms{1.2}{2.8}
& \ms{0.0}{0.0}
& \ms{0.0}{0.0}
& \ms{0.0}{0.0}
& \ms{0.0}{0.0}
& \ms{0.0}{0.0}
& 0.0 \\

GraphIRL
& \ms{34.6}{18.2}
& \ms{3.8}{3.4}
& \ms{0.0}{0.0}
& \ms{0.0}{0.0}
& \ms{0.0}{0.0}
& \ms{0.0}{0.0}
& \ms{0.0}{0.0}
& 0.0 \\

RoboHorizon$^\dagger$
& -
& 75.2
& 74.8
& 48.2
& 31.2
& 36.5
& 40.5
& 39.1 \\

\rowcolor{gray!20}
\textbf{Ours}
& \bms{91.2}{11.8}
& \ms{92.5}{16.8}
& \ms{98.8}{2.8}
& \bms{95.0}{11.2}
& \bms{81.5}{12.0}
& \bms{56.2}{33.1}
& \bms{64.8}{9.6}
& \textbf{74.4} {\color{green!50!black} \(\mathbf{(\Delta 25.4)}\)} \\

\bottomrule
\end{tabular}%
}
\vspace{-0.5cm}
\end{table*}

%% file: tables/sequential_merging_sr_final.tex
\begin{table*}[t]
\centering
\caption{
Cumulative subtask success rate (\%) on long-horizon tasks, reported as mean $\pm$ std over five seeds.
Values in parentheses indicate conditional success rate (\%) wrt previous completed stage. Last column of each task corresponds to full-task success.
\vspace{-0.2cm}
}
\label{tab:sequential_merging_analysis}
\setlength{\tabcolsep}{3.5pt}
\renewcommand{\arraystretch}{1.1}
\resizebox{\textwidth}{!}{%
\begin{tabular}{lcc ccc ccc ccc}
\toprule
Method
& \multicolumn{2}{c}{PutItemInDrawer}
& \multicolumn{3}{c}{PutShoesInBox}
& \multicolumn{3}{c}{PutShoesOutBox}
& \multicolumn{3}{c}{EmptyContainer} \\
\cmidrule(lr){2-3}
\cmidrule(lr){4-6}
\cmidrule(lr){7-9}
\cmidrule(lr){10-12}
& S1 & S1$\rightarrow$S2
& S1 & S1$\rightarrow$S2 & S1$\rightarrow$S2$\rightarrow$S3
& S1 & S1$\rightarrow$S2 & S1$\rightarrow$S2$\rightarrow$S3
& S1 & S1$\rightarrow$S2 & S1$\rightarrow$S2$\rightarrow$S3 \\
\midrule

\rowcolor{gray!10}
Env. Rew.
& \bms{100.0}{0.0} & \ms{0.0}{0.0} \cond{0.0}
& \ms{96.2}{8.4} & \ms{0.0}{0.0} \cond{0.0} & \ms{0.0}{0.0} \cond{0.0}
& \bms{92.0}{17.9} & \ms{0.0}{0.0} \cond{0.0} & \ms{0.0}{0.0} \cond{0.0}
& \ms{20.0}{25.5} & \ms{0.0}{0.0} \cond{0.0} & \ms{0.0}{0.0} \cond{0.0} \\

\rowcolor{gray!10}
Env. Rew. (decomp.)
& \bms{100.0}{0.0} & \ms{62.5}{39.9} \cond{62.5}
& \bms{99.4}{1.3} & \ms{73.8}{41.8} \cond{74.2} & \ms{68.7}{39.0} \cond{93.0}
& \ms{60.0}{54.8} & \ms{60.0}{54.8} \cond{100.0} & \ms{41.2}{39.9} \cond{68.8}
& \bms{98.8}{2.8} & \ms{68.7}{17.7} \cond{69.6} & \ms{23.8}{12.0} \cond{34.6} \\

\midrule

XIRL
& \ms{0.0}{0.0} & \ms{0.0}{0.0} \cond{0.0}
& \ms{66.2}{41.4} & \ms{0.0}{0.0} \cond{0.0} & \ms{0.0}{0.0} \cond{0.0}
& \ms{0.0}{0.0} & \ms{0.0}{0.0} \cond{0.0} & \ms{0.0}{0.0} \cond{0.0}
& \ms{0.0}{0.0} & \ms{0.0}{0.0} \cond{0.0} & \ms{0.0}{0.0} \cond{0.0} \\

GraphIRL
& \ms{0.0}{0.0} & \ms{0.0}{0.0} \cond{0.0}
& \ms{60.0}{22.8} & \ms{0.0}{0.0} \cond{0.0} & \ms{0.0}{0.0} \cond{0.0}
& \ms{3.8}{3.5} & \ms{0.0}{0.0} \cond{0.0} & \ms{0.0}{0.0} \cond{0.0}
& \ms{0.0}{0.0} & \ms{0.0}{0.0} \cond{0.0} & \ms{0.0}{0.0} \cond{0.0} \\

\rowcolor{gray!20}
\textbf{Ours}
& \bms{100.0}{0.0} & \bms{95.0}{11.2} \cond{95.0}
& \ms{96.0}{5.6} & \bms{88.7}{8.7} \cond{90.4} & \bms{81.5}{12.0} \cond{94.0}
& \ms{80.0}{44.7} & \bms{73.8}{41.8} \cond{92.2} & \bms{56.2}{33.1} \cond{76.3}
& \ms{96.2}{5.6} & \bms{70.0}{8.2} \cond{72.7} & \bms{64.8}{9.6} \cond{92.6} \\

\bottomrule
\end{tabular}%
}
\vspace{-0.6cm}
\end{table*}

%% file: tables/reward_discrimination.tex
\begin{table}[t]
\centering
\caption{Reward discrimination on \emph{PutShoesInBox} under standard validation, appearance shift, and detection noise. Lower is better. \vspace{-0.2cm}}
\label{tab:shoes_ratio}
\small
\setlength{\tabcolsep}{3.5pt}
\resizebox{\columnwidth}{!}{%
\begin{tabular}{lccc}
\toprule
Method & Standard $\rho$ $\downarrow$ & Color shift $\rho$ $\downarrow$ & Detection noise $\rho$ $\downarrow$ \\
\midrule
XIRL & 0.898 & 1.759 & - \\
GraphIRL (robot included) & 0.794 & 0.794 & 0.785 \\
GraphIRL (robot removed) & 0.877 & 0.877 & 0.876 \\
Ours (robot included) & 0.760 & 0.760 & 0.985 \\
Ours (robot masked) & \textbf{0.629} & \textbf{0.629} & \textbf{0.583} \\
\bottomrule
\end{tabular}%
}
\vspace{-10pt}
\end{table}

%% file: tables/ablation_bonus.tex
\begin{table}[t]
\centering
\caption{Success rate (\%) on different steps of \emph{PutShoesInBox}. 
The composed-policy column refers to sequential composition of the learned subtask policies. Higher is better.\vspace{-0.2cm}}
\label{tab:multi_task_success}
\resizebox{\columnwidth}{!}{%
\begin{tabular}{lcccc}
\toprule
Setting & OpenBox & PlaceFirstShoe & PlaceSecondShoe & Composed policy \\
\midrule
\rowcolor{gray!15}
Env. bonus only 
& \ms{98.8}{2.6} 
& \ms{75.0}{43.3} 
& \ms{0.0}{0.0} 
& \ms{0.0}{0.0} \\

\rowcolor{gray!15}
Env. reward  
& \bms{100.0}{0.0} 
& \ms{80.0}{44.7} 
& \ms{80.0}{44.7} 
& \ms{68.7}{39.0} \\

Ours + no bonus 
& \ms{86.4}{12.7} 
& \ms{2.5}{3.4} 
& \ms{0.0}{0.0} 
& \ms{0.0}{0.0} \\

Ours + env bonus 
& \bms{100.0}{0.0} 
& \ms{80.0}{44.7} 
& \ms{80.0}{44.7} 
& \ms{58.0}{52.9} \\

Ours + learned bonus 
& \ms{98.8}{2.7} 
& \bms{98.8}{2.7} 
& \bms{95.0}{11.2} 
& \bms{81.5}{12.0} \\

\bottomrule
\end{tabular}%
}
\vspace{-0.4cm}
\end{table}

%% file: tables/ablation_loss_shoes.tex
\begin{table}[t]
\centering

\caption{Training objective ablation on \emph{PutShoesInBox}. Lower is better.\vspace{-0.2cm}}
\label{tab:ablation_losses}
\small
\setlength{\tabcolsep}{4pt}
\begin{tabular}{lc}
\toprule
Method & Ratio $\rho$ $\downarrow$ \\
\midrule
TCC loss only & 0.798 \\
Reconstruction loss only & 0.837 \\
Ours (both) & \textbf{0.629} \\
\bottomrule
\end{tabular}%
\vspace{-10pt}
\end{table}

%% file: sections/05.discussion.tex
The results support the central hypothesis of this work: our graph-based object-centric representations provide a useful abstraction for learning rewards from action-free demonstrations in robotic manipulation. 
Compared with pixel-level and simpler object-centric reward methods, our representation reduces dependence on low-level appearance cues while preserving the entities and spatial relations that define manipulation success. 
In long-horizon tasks, the learned reward profiles show that the proposed representation captures ordered object-state transitions as stage-like changes, without subtask annotations.

In short-horizon and single-stage tasks, the full-task learned reward can be used directly as a dense shaping signal for reinforcement learning and is sufficient to guide policy optimization. 
In long-horizon tasks, however, using a single full-task reward is generally not enough: the reward must also provide information about the internal stage structure of the task. 
The step-like reward profiles observed across long-horizon tasks suggest that the learned latent distance to the goal is consistent with sensitivity to semantic progress events rather than only final-frame similarity.
This property makes the full-task reward model a natural basis for automatic subtask discovery, after which subtask-specific rewards can be used to train specialized policies.

The long-horizon results also highlight an important distinction between partial progress and full task completion. 
Some baselines, including the single full-task environmental reward, XIRL, and GraphIRL, are able to solve early stages of certain tasks, such as opening a drawer or opening a box. 
However, their performance drops sharply in later cumulative stages. 
This suggests that the difficulty is not only learning any useful reward signal, but maintaining a reward structure that remains informative across multiple ordered phases. 
By discovering subtask boundaries and training localized rewards for each phase, our method provides more targeted supervision for sequential policy learning.

%% file: sections/06.conclusions.tex
We presented GORDON, a graph-based object-centric reward learning framework that learns rewards for robotic manipulation from action-free video demonstrations. 
The method represents observations as graphs of detected objects and spatial relations, and uses activity-aware weighted pooling to emphasize task-relevant object changes while masking robot-dominated motion, producing reward signals that are more semantically aligned with the task.

Experiments across seven manipulation tasks show that the learned rewards transfer effectively to reinforcement learning. 
In short-horizon and single-stage settings, the reward can be used directly as a dense shaping signal for policy optimization. 
In long-horizon tasks, its stage-wise temporal profile enables automatic subtask discovery, subtask-specific reward learning, and sequential composition of specialized policies.

Overall, these results suggest that our learned rewards can both guide policy optimization and expose semantic task structure that simplifies long-horizon manipulation.